\documentclass[a4paper, 10pt, conference]{ieeeconf}      

\IEEEoverridecommandlockouts                              
\usepackage{graphics} 
\usepackage{graphicx}
\usepackage{caption}
\usepackage{epsfig} 
\usepackage{times} 
\usepackage{amsmath} 
\usepackage{amssymb}  
\usepackage{subfig}
\usepackage{nameref}
\usepackage{todonotes}

\def\usenatbib{1}
\ifx\usenatbib\undefined
    \usepackage{cite}
\else
    \makeatletter
    \let\NAT@parse\undefined
    \makeatother
    \usepackage[numbers,sort]{natbib}
    \setcitestyle{citesep={], [}}
    \makeatletter
    \def\NAT@def@citea{\def\@citea{\NAT@separator}}%
    \makeatother
\fi
\usepackage[colorlinks=true,linkcolor=black,anchorcolor=black,citecolor=black,filecolor=black,menucolor=black,runcolor=black,urlcolor=black]{hyperref}

\let\orgautoref\autoref
\providecommand{\Autoref}
        {\def\equationautorefname{Equation}%
         \def\figureautorefname{Figure}%
         \def\subfigureautorefname{Figure}%
         \def\Itemautorefname{Item}%
         \def\tableautorefname{Table}%
         \def\exerciseautorefname{Exercise}%
         \def\starexerciseautorefname{Exercise}%
         \def\sectionautorefname{Section}%
         \def\subsectionautorefname{Section}%
         \def\subsubsectionautorefname{Section}%
         \def\chapterautorefname{Section}%
         \def\partautorefname{Part}%
         \orgautoref}

\renewcommand{\autoref}
        {\def\equationautorefname{Equation}%
         \def\figureautorefname{Fig.}%
         \def\subfigureautorefname{Fig.}%
         \def\Itemautorefname{item}%
         \def\tableautorefname{Table}%
         \def\exerciseautorefname{Exercise}%
         \def\starexerciseautorefname{Exercise}%
         \def\sectionautorefname{Section}%
         \def\subsectionautorefname{Section}%
         \def\subsubsectionautorefname{Section}%
         \def\chapterautorefname{Section}%
         \def\partautorefname{Part}%
         \orgautoref}

\DeclareMathOperator{\RMSE}{RMSE}
\DeclareMathOperator{\sgn}{sgn}

\graphicspath{{figures/}}

\title{\LARGE \bf
Adaptive Selection of Informative Path Planning Strategies 
via Reinforcement Learning
}

\author{Taeyeong Choi$^{1}$ and Grzegorz Cielniak$^{1}$
\thanks{$^{1}$Taeyeong Choi and Grzegorz Cielniak are with Lincoln Institute 
        for Agri-food Technology, University of Lincoln, Riseholme Park, 
        LN2 2LG Lincoln, UK
        {\tt\small \{tchoi, gcielniak\}@lincoln.ac.uk}
        \newline 978-1-6654-1213-1/21/\$31.00 \textcopyright 2021 IEEE}%
}

\begin{document}
\maketitle
\thispagestyle{empty}
\pagestyle{empty}

\begin{abstract}

In our previous work, we designed a systematic policy to prioritize 
sampling locations to lead significant accuracy improvement in 
spatial interpolation by using the prediction uncertainty of 
\emph{Gaussian Process Regression}~(GPR) as ``attraction force'' to 
deployed robots in path planning.
Although the integration with Traveling Salesman Problem~(TSP) solvers 
was also shown to produce relatively short travel distance, we here 
hypothesise several factors that could decrease the overall prediction precision as well 
because sub-optimal locations may eventually be included in their paths. 
To address this issue, in this paper, we first explore
``local planning'' approaches adopting various spatial ranges 
within which next sampling locations are prioritized to 
investigate their effects on the prediction performance as well 
as incurred travel distance. 
Also, Reinforcement Learning~(RL)-based \emph{high-level}
controllers are trained to \emph{adaptively} produce blended plans from a particular 
set of local planners to inherit unique strengths from that selection 
depending on latest prediction states.  
Our experiments on use cases of temperature monitoring robots 
demonstrate that the dynamic mixtures of planners can not only generate 
sophisticated, informative plans that a single planner could not create alone 
but also ensure significantly reduced travel distances at no cost of prediction 
reliability without any assist of additional modules for shortest path calculation.

\end{abstract}

\section{Introduction}
\label{sec:introduction}

Mobile robotic agents can act as very useful sensory instruments for automatic monitoring of spatio-temporal phenomena, especially when the field of interest is too large or too risky for humans to explore, or when the frequent updates are required due to continuous changes over time. Therefore, a wide range of potential applications have been introduced to utilise autonomous samplers offering precise estimation of spatial attributes such as air temperature~\citep{LS18}, 
plankton density under water~\citep{MLSRD18}, or 
compaction or moisture level in soil~\citep{FGDPC18, PBDEPC20}.   
In realistic scenarios, however, robots also have some limitations 
related to their physical properties (e.g., velocity) and resources 
(e.g.,~battery life) implying that their navigation plan must be 
strategically designed in the sense of \emph{Informative Path Planning}~(IPP)~\citep{MKGH07} 
to best learn the environmental model by prioritizing gathering of information-rich samples whilst travelling for a limited amount of time~\citep{LS18}. 

\begin{figure}\centering
    \subfloat[]{\label{fig:rl_path1}%
        \includegraphics[width=0.495\columnwidth]{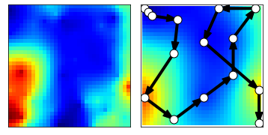}}
    \subfloat[]{\label{fig:rl_path2}%
        \includegraphics[width=0.495\columnwidth]{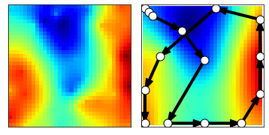}} \\
    \subfloat[]{\label{fig:rl_path2_act}%
        \includegraphics[width=.99\columnwidth]{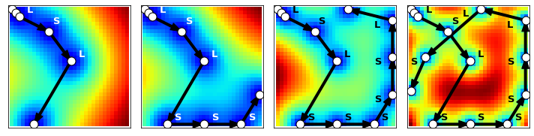}} 
    \caption{
        \protect\subref{fig:rl_path1}, \protect\subref{fig:rl_path2}: 
        Two examples of temperature prediction each showing the ground truth 
        (left) and the GPR interpolation (right) based on
        $15$~samples from the locations~(white circles) proposed by 
        our RL-$21$ method. \protect\subref{fig:rl_path2_act}:
        Kriging variances after $6, 9, 12,$ and $15$~samples 
        in \protect\subref{fig:rl_path2} while \emph{local} planners with 
        ``short'' and ``long'' ranges are dynamically executed. 
        }
    \label{fig:rl_paths}
\end{figure}

In our previous work~\citep{FGDPC18}, we proposed to employ \emph{Ordinary Kriging}~(OK) algorithm ---~a form of 
\emph{Gaussian Process Regression}~(GPR) ---~to not only learn 
the holistic environmental model from the sparse 
measurements of soil compaction but also utilise the estimation 
uncertainty as the ``driving force'' in choosing the next sampling locations.
Though this uncertainty-based planning led to significant improvements in 
overall estimation accuracy as in other similar 
applications~\citep{MLSRD18, CH20}, long travels could inevitably 
be caused by the nature of \emph{Global Search}~(GS), in which  
the prediction uncertainties over the field
were all globally considered for 
planning~(c.f.,~$2$nd in~\autoref{fig:baseline_paths}).

To alleviate this issue, we also suggested reframing the path planning as 
\emph{Traveling Salesman Problem}~(TSP) to regularly compute the shortest 
path through least-confident locations 
($4$th in \autoref{fig:baseline_paths})~\citep{FGDPC18}. 
Yet, here we claim that this modification could not warrant a high degree of reliability 
in prediction, first because some arbitrary points are included for the initial path 
generation, and also because later waypoints may become more certain and less informative while visiting earlier ones. 

In this paper, we thus propose a novel approach for IPP to 
maintain the high precision of interpolation over the explored field 
as well as to significantly reduce the total travel distance.
In particular, we first explore several \emph{Local Search}~(LS) strategies 
with various ``spatial ranges'' only within which highly uncertain locations are determined 
as the next waypoints (e.g.,~$3$rd in~\autoref{fig:baseline_paths}). 
In addition, Reinforcement Learning~(RL)-based \emph{high-level} controllers 
are designed to adaptively switch between the \emph{low-level} LS~planners 
depending on the state of spatial estimation so that the combined execution 
could inherit distinct strengths from the low-level policies 
(\autoref{fig:rl_paths}). 
We also demonstrate practical 
use cases to validate our model in which mobile robots generate informative, 
efficient paths to provide accurate estimates of air quality (e.g.,~temperature or humidity) 
in indoor environment. Lastly, the extensive tests on $120$~instances with real 
temperature readings show that the RL controllers can successfully solve 
IPP by exhibiting dynamic mixtures of plans in adaptive manner as well as 
significantly shorter routes can be planned without loss of prediction precision.


\section{Related Work}
\label{sec:related_work}

\subsection{Informative Path Planning}
\label{sec:ipp}

IPP has been studied over the last decades, and due to its complexity as
a NP-hard problem~\citep{MKGH07}, a number of approaches have been 
proposed to provide approximate optimal paths by considering proxy 
metrics of ``informativeness''. In particular, GPR has been widely adopted to 
build the spatial maps of property online and utilise  
learnt covariance matrices to calculate ``mutual information''
between the visited locations and the rest of the environment~\citep{BKS13, MLS17}.
Also, as in our previous methods~\citep{FGDPC18, PBDEPC20}, 
prediction variance from GPR can be used by planning modules 
to set the least-certain location  as the ``next-best-view'' waypoint 
after each update of the GPR model with a new observation~\citep{MLSRD18}.
Additionally, path planners for TSP can be employed to construct 
the shortest travelling routes for robot navigation~\citep{FGDPC18, MLS17}. 
However, our RL model here is not a substitute for any of these but instead 
focused on learning the best mixtures of such strategies with an assumption that 
novel strengths from different combinations could be revealed in resulting plans 
to better solve IPP.

More recently, \citeauthor{WZ20} in~\citep{WZ20} have also suggested 
using RL for IPP, but their application is limited to speeding up the computation  
of the shortest paths on the locations already chosen by some information-theoretic 
metric. In contrast, ours is designed to draw a new policy from 
multiple strategies with unique advantages.  
Furthermore, the performance of their model has only been reported 
in terms of mutual information from selected paths, while our testbeds 
are directly validated using prediction error with real sensory 
measurements. 

\begin{figure}[t]\centering
    \includegraphics[width=0.95\columnwidth]{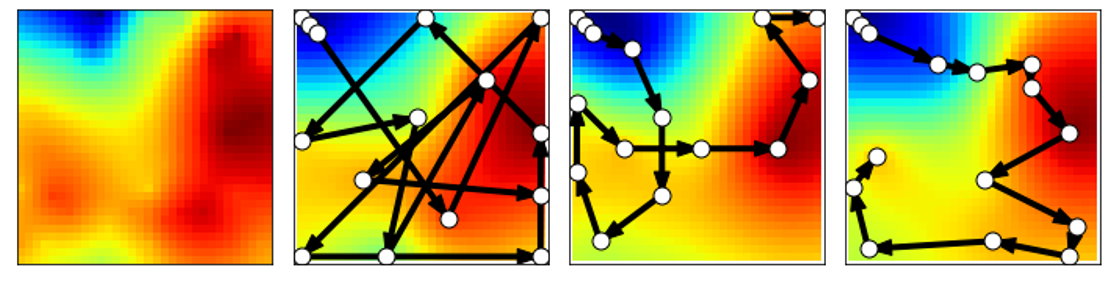}
	\caption{Examples of final prediction outputs with 
    path plans for $15$~samples: Ground truth, GS, LS-1, and GS-TSP. 
    }
	\label{fig:baseline_paths}
\end{figure}

\subsection{Blending Controllers}
\label{sec:blending_controllers}

In the context of IPP, \citeauthor{MLS17} have formulated 
a ``soft-blending'' function \citep{MLS17} in which each candidate location 
is evaluated with weighted summation of the travel distance and GPR's 
estimation variance to balance between the final performance in precision 
and efficiency. In~\citep{GDVT20}, a more relevant model, outside IPP, was 
built with multi-armed bandits upon two types of specialised robotic controllers 
---~one of high performance and one of high safety ---~for mobile 
robots to best perform the collision avoidance in autonomous navigation. 
The main distinction in our work is that the low-level specialised controllers 
are not the expensive products of RL, and instead our goal is to combine relatively 
\emph{cheap} planners to produce a sophisticated, complementary plan that could 
not be generated by simply using any of them alone.

\section{Problem Description}
\label{sec:problem_description}

In this section, we first formally describe IPP problem, and then 
introduce a specific use case scenario of robots monitoring temperature which will also be tested with datasets collected from real sensor measurements. 

\subsection{Informative Path Planning}
\label{sec:informative_path_planning}

We consider a \emph{path planning} problem for a mobile robot~$\rho$ 
deployed in a discretised two-dimensional 
field~$\mathcal{F} \subseteq \mathbb{R}^2$  of $N$ grid 
locations, indexed as $\ell \in \{1,2,...,N\}$,
to estimate an unknown density function $\mathcal{Z}$ returning
the environmental attribute~$z_\ell \in \mathbb{R}$
associated with location~$\ell$. 
In particular, we can assume that $\mathcal{F}$ is obstacle-free 
space, and robot~$\rho$ must physically \emph{visit} a specific 
location~$\ell$ to assess $z_{\ell} = \mathcal{Z}(\ell)$.

In general, the objective in IPP is to discover the optimal
policy~$\pi^{*}$ to generate a path 
$\mathcal{L}_{T}=( \ell_1, \ell_2, ..., \ell_T )$ 
at discrete timesteps $1 \leq t \leq T$ and use the observations 
$\zeta_{T}=(z_{\ell_1}, z_{\ell_2}, ..., z_{\ell_T})$ 
 to minimise the prediction error of interpolator~$f$, which can 
be evaluated by the Root Mean-Squared Error~(RMSE):

\begin{equation}
    \RMSE \big( \hat{\mathcal{Z}}_{T}, \mathcal{Z} \big)
    = \sqrt{\frac{1}{N} \sum_{\ell=1}^{N} (\hat{z}_{\ell} - z_{\ell})^{2}},
    \label{eq:rmse}
\end{equation}
where $\hat{\mathcal{Z}}_{T}$ is the prediction of function~$\mathcal{Z}$ 
from interpolator~$f(\zeta_{T}; \theta_{T})$. 
Note here that $f$ can be parameterized by~$\theta_T$ learnt  
from the observations~$\zeta_T$, and thus, IPP can be also seen 
as \emph{active learning}~\citep{FLC17}, in which $\pi^*$ is discovered 
to choose the optimal stream of observations to learn $\theta^*$ 
that can best model the true environmental function~$\mathcal{Z}$. Similarly,
our novel strategy is to approximate $\pi^*$ by selecting among a set 
of suboptimal policies~$\{\pi_1, \pi_2, ..., \pi_k \}$ at each time of planning 
for $\ell_t$ so that the resulting sequential observations~$\zeta_t$ can 
minimise the final prediction error in~\autoref{eq:rmse}.

\subsection{Use Case: Temperature Monitoring Robots}
\label{sec:use_case}

As an application scenario, we propose to deploy a mobile robot for autonomous monitoring of air quality ---~for example, 
temperature or humidity. 
For realistic testbeds,  the Intel Berkeley Lab 
dataset\footnote{http://db.csail.mit.edu/labdata/labdata.html} is used 
as in~\citep{MKGH07, LS18, LNKS19}, in which $54$~sensors were placed in an 
indoor office to collect values of spatio-temporal air quality once every $31$~seconds for over a month in $2004$. 
In particular, we utilise the data of temperature as a representative 
attribute~$\mathcal{Z}$ to explore, ignoring the readings from four sensors 
which feature faulty or missing values. Similarly 
to~\citep{FGDPC18, PBDEPC20}, spatial interpolation using Kriging
(c.f.~\autoref{sec:ordinary_kriging}) was also performed on the remaining values, 
and then an image-rescaling technique was applied to gain $(32 \times 32)$  
dense ``surrogate'' instances as realistic ground truth data as shown 
in~\autoref{fig:rl_paths} and~\autoref{fig:baseline_paths}. 
Finally, we randomly chose $240$~instances with an exclusive split of $120$ 
and $120$ for training and test, respectively. 

\section{Uncertainty-driven Planning}
\label{sec:uncertainty_driven_planning}

Here, we explain OK interpolation techniques and our previous planning method 
based upon them. In addition, several potential limitations are discussed to 
present motivation for our proposed approach, introduced in~\autoref{sec:proposed_methodology}.

\subsection{Ordinary Kriging for Spatial Interpolation}
\label{sec:ordinary_kriging}

As interpolator~$f$, we employ 
Ordinary Kriging~(OK), which is a GPR technique in which prior 
knowledge of global mean is not required, whereas 
na\"ive GPR, analogous to Simple Kriging~(SK), assumes the zero 
mean or utilises the sample mean from the available 
training data~\citep{L13}. 
For a location~$\ell_{0}$, the OK-based prediction 
is performed by the linear combination 
of~$z_{\ell_{i}}$ measured at previous steps~$i=1, ..., t$: 

\begin{equation}
    \hat{\mathcal{Z}}_{t}(\ell_{0}) := \\
    \sum_{i=1}^{t} w_{i} z_{\ell_{i}}, 
    \label{eq:kriging_wz}
\end{equation}
where $w_i \in \mathbb{R}$ is essentially designed to depend 
on the distance to~$\ell_{i}$, and $\sum_{i=1}^{t} w_{i}=1$ 
to ensure the unbiased estimator. Also, 
$\mathbf{w} = [w_{1}, w_{2}, ..., w_{t}]^{T}$
can be gained by solving the following system in matrix 
formulation~\citep{L13}:

\begin{equation}
    \begin{bmatrix}
        \mathbf{w}\\
        \lambda
    \end{bmatrix}
    =
    \begin{bmatrix}
        \mathbf{\Gamma}_{ij} & \mathbf{1}\\
        \mathbf{1}^{T} & 0
    \end{bmatrix}^{-1}
    \begin{bmatrix}
        \mathbf{\Gamma}_{i0}\\
        1
    \end{bmatrix},
    \label{eq:kriging_matrix}
\end{equation}
where $\lambda$~is the Lagrange multiplier, 
$\mathbf{\Gamma}_{i0} = [\gamma(h_{10}), \gamma(h_{20}), ..., \gamma(h_{t0})]^T$
with the \emph{variogram} function~$\gamma$, which takes the distance
between $\ell_i$ and $\ell_0$ as input, and similarly, 
$\mathbf{\Gamma}_{ij}$ is the symmetric variogram matrix for 
all possible pairs of visited locations~$(\ell_i, \ell_{j})$ 
--- i.e.,~$\mathbf{\Gamma}$ plays a similar role of the covariance 
matrix in basic GPR models. Specifically, we adopt the \emph{spherical} model of 
variogram function~$\gamma$ to 
produce~$p(\frac{3h}{2r} - \frac{h^3}{2r^3})+n$
if $h \leq r$, and $p+n$ otherwise, where $h$ is the input distance, and 
$p, r,$ and $n$ are the learnt parameters in~$f$ called \emph{partial sill, range,} 
and \emph{nugget}, respectively, based on the historical measurements~$\zeta_t$. 
In this work, we keep three locations 
with $xy$-coordinates $((1,1),(2,2),(3,3))$ as a \emph{seed} 
set~$\mathcal{L}_3$ for OK-based methods to perform initial 
estimation of those parameters.

\subsection{Global Search for Uncertain Locations}
\label{sec:global_search_for_uncertain_locations}

In~\citep{FGDPC18}, our GS approach was to build a 
``next-best-view'' planning policy~$\pi$ utilising the estimated 
prediction variance~$\sigma_{\ell_0}^2$, called \emph{Kriging Variance}~(KV), 
from true target value to quantify the 
uncertainty of each prediction outcome. More specifically, 
once the OK system in~$\autoref{eq:kriging_matrix}$ has been solved, 
the KV for~each $\ell_0$ can easily be calculated as follows: 
$\sigma_{\ell_0}^2 = \lambda + \mathbf{w}^{T} \Gamma_{i0}$.
Therefore, $\pi$ evaluates this quantity over the entire field to \emph{guide} 
robots to the location~$\ell^{KV}$ of its highest level
because in~\citep{FGDPC18}, we discovered a strong correlation between 
the KV at sampling locations and the reduction amount of interpolation error. 
However, note here that in this work, particularly, we instead utilise 
\emph{noisy} KV, i.e.,~$\sigma_{\ell_0}^2 + b$ where 
$b \sim N(0, 10^{-12})$, to induce some randomness in $\pi$, which 
we have found more useful for robots to be more exploratory 
to improve prediction accuracy.

\subsection{Integration with TSP Solvers \& Potential Limitations}
\label{sec:integration_with_tsp_solvers}

Albeit KV was an useful indicator in our previous GS~model, 
impractical paths with overly long travel distances tended 
to be generated as in~\autoref{fig:baseline_paths} because all locations 
over the field were considered in comparison of KV's. 
To tackle this issue, we further suggested using TSP solvers 
as an additional module in~\citep{FGDPC18} to compute shortest paths whenever a 
newly sampled observation was acquired. In fact, this integration was shown to 
assist in producing shorter travel distances, but we here hypothesize that 
there may exist several critical factors in the TSP's involvement which might 
significantly \emph{degrade} overall prediction reliability.

\subsubsection{TSP random samples} 
The TSP module requires arbitrary locations~$\ell_{rand}$ to be drawn in the queue 
for initial path planning, and these could remain to eventually be sampled as 
next waypoints for robots to visit, although a high degree of informativeness 
cannot be ensured from those locations. 

\subsubsection{Delays until visit}
The visit to a location~$\ell$ in the planning 
queue often occurs some steps after it was actually added  
since in the meantime, closer locations~$\ell'$ to the robot are visited
first. Consequently, $\ell$ may not have the largest KV at the timing
of visit because the earlier observations from~$\ell'$ may have changed
prediction certainty over the field. 

To validate these hypotheses, extensive evaluations are performed 
in~\autoref{sec:problem_in_GS_TSP}.



\begin{figure}\centering
    \subfloat[]{\label{fig:tradeoff}%
        \includegraphics[width=0.495\columnwidth]{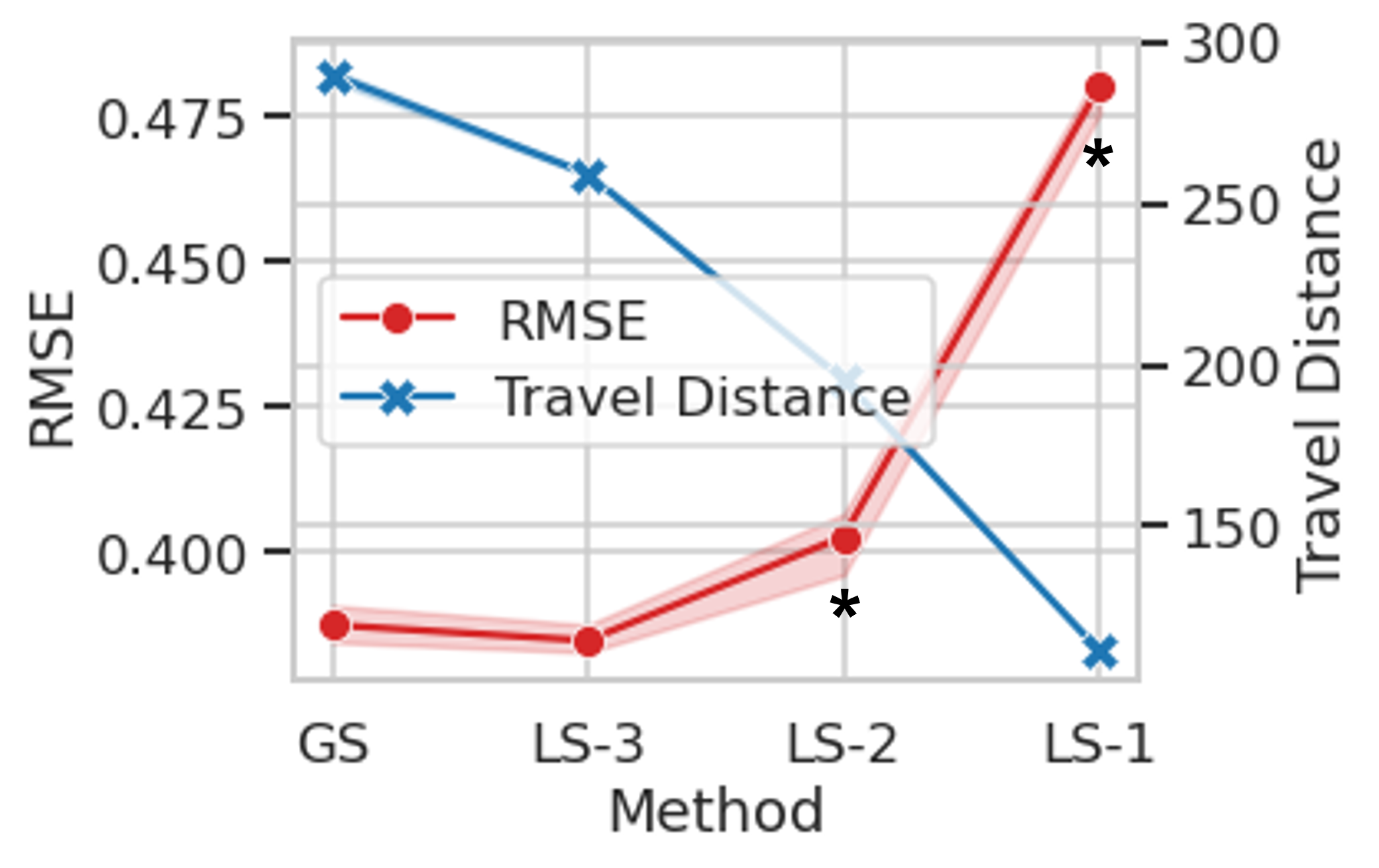}}
    \subfloat[]{\label{fig:rl_learning}%
        \includegraphics[width=0.495\columnwidth]{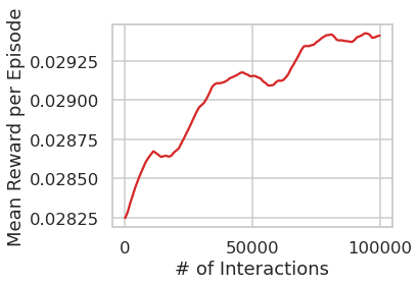}}
    \caption{
        \protect\subref{fig:tradeoff}: Prediction errors and total 
        travel distance incurred by GS and LS after 
        $15$th sample. ``*'' marks a significant increase of error 
        compared to GS ($p < .05)$;  
        \protect\subref{fig:rl_learning}: Increasing trend of the mean 
        reward per episode over training in RL. A rolling average 
        has been applied to mitigate noise for clarity.         
        }
    \label{fig:tradeoff_rl_learning}
\end{figure}

\section{Proposed Methodology}
\label{sec:proposed_methodology}

In this section, we explain core components of our novel framework 
to ultimately generate efficient but most informative paths for robotic sampling. 

\subsection{Local-Search Strategies}
\label{sec:local_search_strategies}

We perform here closer investigations on the behavior of the 
next-best-view policy~$\pi$ with some modifications --- i.e.,~inspired 
by ``fixed-window location selection'' in~\citep{MLSRD18}, we build 
LS~policies to \emph{locally} search for the next locations only 
within ``preset distances'' as displayed in~\autoref{fig:baseline_paths}. 

\subsubsection{Reliability vs Efficiency}
\label{sec:reliability_vs_efficiency}

 \Autoref{fig:tradeoff} visualizes the average estimation error 
and travel distance after $15$th sample when the LS
models such as LS-$1$, LS-$2$, and LS-$3$ with ranges 
of $10, 20,$ and $30$, respectively, are compared with our previous 
GS model to clearly reveal: 
(1)~Robots with wider coverages tend to produce lower errors, 
(2)~Smaller ranges more likely incur shorter travel distances, and as a result, 
(3)~The \emph{trade-off} between the two metrics may imply that minimising 
one would necessarily maximise the other when utilising the KV-based models.

\subsubsection{LS-winning Examples}
\label{sec:ls_winning_examples}

Although the GS method can achieve significantly more accurate predictions 
than short LS models (e.g.,~LS-$1$ and LS-$2$), we also have discovered  
field instances in which the local searches can outperform GS in precision. 
To be specific, LS-$1$ wins in $27$~($23\%$) out of $120$~test fields, and LS-$2$ also 
does in $42$~($35\%$) instances implying that global-maximum
KV can generally be a ``good'' indicator used for guidance but can be a 
nonoptimal choice in some conditions. LS-$3$'s similar performance to GS 
may also support this intuition. In other words, if short- and long-range 
planners are \emph{collaboratively} used at effective times, the estimation error 
and the travel cost both could be minimised simultaneously taking advantage of 
their unique strengths ---~an \emph{ideal} policy to overcome the 
trade-off discussed above.

Motivated by these observations, we design a \emph{higher-level} controller 
to decide which strategy to use for planning by considering contextual information. 

\subsection{Mixture of Plans via Reinforcement Learning}
\label{sec:reinforcement_learning}

\subsubsection{Markov Decision Process}
RL problems can be represented by a Markov Decision Process~(MDP), in which 
at each discrete time step~$t$, the RL agent~$\rho$ uses its policy function~$\Pi_\theta$ 
to select an action~$a_t$ from environmental state~$s_t$ and receive a scalar reward 
signal~$r_t$ which would reflect the performance of the agent in the task. 
The aim of RL is to discover the optimal policy~$\Pi_{\theta^{*}}$ to maximise the expected 
cumulative reward from time instant~$t$~\citep{GK19}: 
$R_{t} =
\mathbb{E}_{\tau \sim \Pi_\theta} \big[\sum_{t'=t}^{T} \gamma^{t'-t}r_{t'} \big]$
where $\gamma \in [0, 1]$ is a discount factor to more 
prioritize immediate rewards with $\gamma=0$ or treat all rewards equally with 
$\gamma=1$. The expectation is not directly accessible in practical 
scenarios, in which the transition function~$\mathcal{P}$ mapping from 
$(s_t, a_t)$ to $s_{(t+1)}$ is unknown, and thus, approximation is 
performed by sampling a number of trajectories 
$\tau=(s_t, a_t, r_t, ..., s_T, a_T, r_T)$ by repeated execution 
of~$\Pi_\theta$.

In our particular case, the state $s_t$ after $(t-1)$th sample is defined as 
the tuple of the last Kriging-prediction mean and variance
$M_{t-1},V_{t-1} \in \mathbb{R}^{H\times W}$, 
the current position of robot
$P_{t} \in \mathbb{R}^{H\times W}$, which is a zero matrix with a value 
of $1$ only at the current location, and the normalized scalar of current 
sampling step~$t'=t/T$ where $T$ is the targeted number of samples in budget.
$\Pi_\theta$ selects its action 
policy~$a_t \in \{\pi_{1}, \pi_{2}, ..., \pi_{k} \}$ to execute and 
gather in history~$\zeta_{t}$ the corresponding 
observation~$z_{t}=\mathcal{Z}(\ell_{t})$ from the 
location $\ell_{t}=a_t(s_t)$ in order to ultimately predict 
$\hat{\mathcal{Z}}_{t}$.

For calculation of reward signal~$r_t$, our approach is first 
to evaluate improved performance~$\Delta^a_t$ by taking action~$a$ over 
any other hypothetical actions~$\bar{a}$ 
(so-called ``hallucination'' in~\citep{CP20}):

\begin{equation}
    \Delta^a_t = 
    \min_{\bar{a} \in \{\pi_1,...,\pi_k\} \setminus \{a\}} 
    \RMSE\big(\hat{\mathcal{Z}}^{\bar{a}}_{t}, \mathcal{Z} \big) -
    \RMSE\big(\hat{\mathcal{Z}}_{t}, \mathcal{Z} \big).
\end{equation}
Then, the reward~$r_t$ is obtained as follows:


\begin{equation}
    r_{t} = 
    \frac{t'}{\big(C-\RMSE(\hat{\mathcal{Z}}_{t}, \mathcal{Z})\big)^{\beta}} 
    \sgn(\Delta^a_t),
\label{reward_function}    
\end{equation}
where $C$ is an arbitrary constant large enough to keep the denominator positive,
$\sgn(x)$ is set to return $1$ if $x\geq 0$ and $-1$ otherwise, 
$t'=t/T$ to present higher rewards for low errors at later visits, and 
$\beta$ is a hyperparameter to control the weight on the estimation error.
Basically, $\Pi_\theta$ is punished unless the chosen action~$a$ 
brings about the lowest prediction error among others, and  
the magnitude of signal~$r_t$ can be determined by the prediction 
error. Yet, $t'$ would generally allow for risks of high error at earlier 
sampling steps to minimise the final error after~$T$.
From our empirical results, we set $C=1$ and $\beta=4e$ for 
stable learning progress.

\subsubsection{Optimization with DQN}

For optimization purpose, $R_t$ can be described in the form of 
$Q\text{-}value$ function with direct relation to states and 
actions in~$\Pi_\theta$~\citep{GK19}:
    $Q_{\Pi_\theta}(s,a) = \mathbb{E}_{\tau\sim\Pi_\theta}
    \big[ R_t \big| s, a \big]$
%
where $Q_{\Pi_\theta}$ specifies a particular state~$s$ at~$t$
from which the expected cumulative reward is evaluated by first taking 
action~$a$ and following $\Pi_\theta$ afterwards. The optimal $Q$~function
can also be described by \emph{Bellman equation} as follows~\citep{MKSGAWR13}:

\begin{equation}
    Q^{*}_{\Pi_\theta}(s,a) = \mathbb{E}_{s'\sim \mathcal{P}}
    \bigg[r + \gamma \max_{a'}Q^*(s', a') \bigg| s, a \bigg],
    \label{eq:bellman}
\end{equation}
where $s'$ and $a'$ are the state and the action at the next step, 
respectively, and especially, $s'$ is determined by the transition 
function~$\mathcal{P}$ in the environment. 
As in~\citep{MKSGAWR13}, we also adopt ``Deep Q-Networks''~(DQN)
with parameters~$\theta$ as a function approximator of $Q^*$, and thus, 
the following loss function is used to minimise at each iteration~$i$: 

\begin{equation}
    \mathcal{L}(\theta_i) = 
    \mathbb{E}_{s, a\sim \Gamma}
    \bigg[ (y_i - Q(s, a; \theta_i))^2 \bigg],
    \label{eq:loss}
\end{equation}
where $y_i = \mathbb{E}_{s'\sim \mathcal{P}}
\big[ r+\gamma max_{a'} Q(s', a'; \theta_{i-1}) \big| s,a \big]$, and
$\Gamma(s,a)$ is a probability distribution over encountered sequences of 
states and actions, which is approximated via a number of interactions 
in trajectory~$\tau$.

\subsection{Neural Network Architectures}
\label{sec:neural_network_architectures}

Since our input state~$s_t$ contains a set of two dimensional vectors
$\{M_{t-1}, V_{t-1}, P_t\}$, four convolutional layers with 
$64, 128, 256,$ and $256$ $(3\times3)$ 
filters are mainly deployed in our DQN to extract useful spatial features. 
Each output from convolutions is downsampled by max pooling operations to 
finally produce the flattened output~$v_1 \in \mathbb{R}^{1024}$ from the 
last convolutional layer. Furthermore, the input scalar~$t'$ is processed 
by a dense layer with $8$~nodes to compute a vector representation $v_2 \in \mathbb{R}^8$, 
and then $v_1$ and $v_2$ are concatenated as an input to two serial dense 
layers with $1024$~and $|\mathcal{A}|$~nodes, respectively, where 
$|\mathcal{A}|$ is the number of actions to learn $Q$~values for 
corresponding actions in the output layer. 
Every layer employs the \emph{LeakyReLU} function 
for activation except the last output layer with linear activation. 

\begin{figure}\centering
    \subfloat[Prediction Error]{\label{fig:prediction_error}%
        \includegraphics[width=0.495\columnwidth]{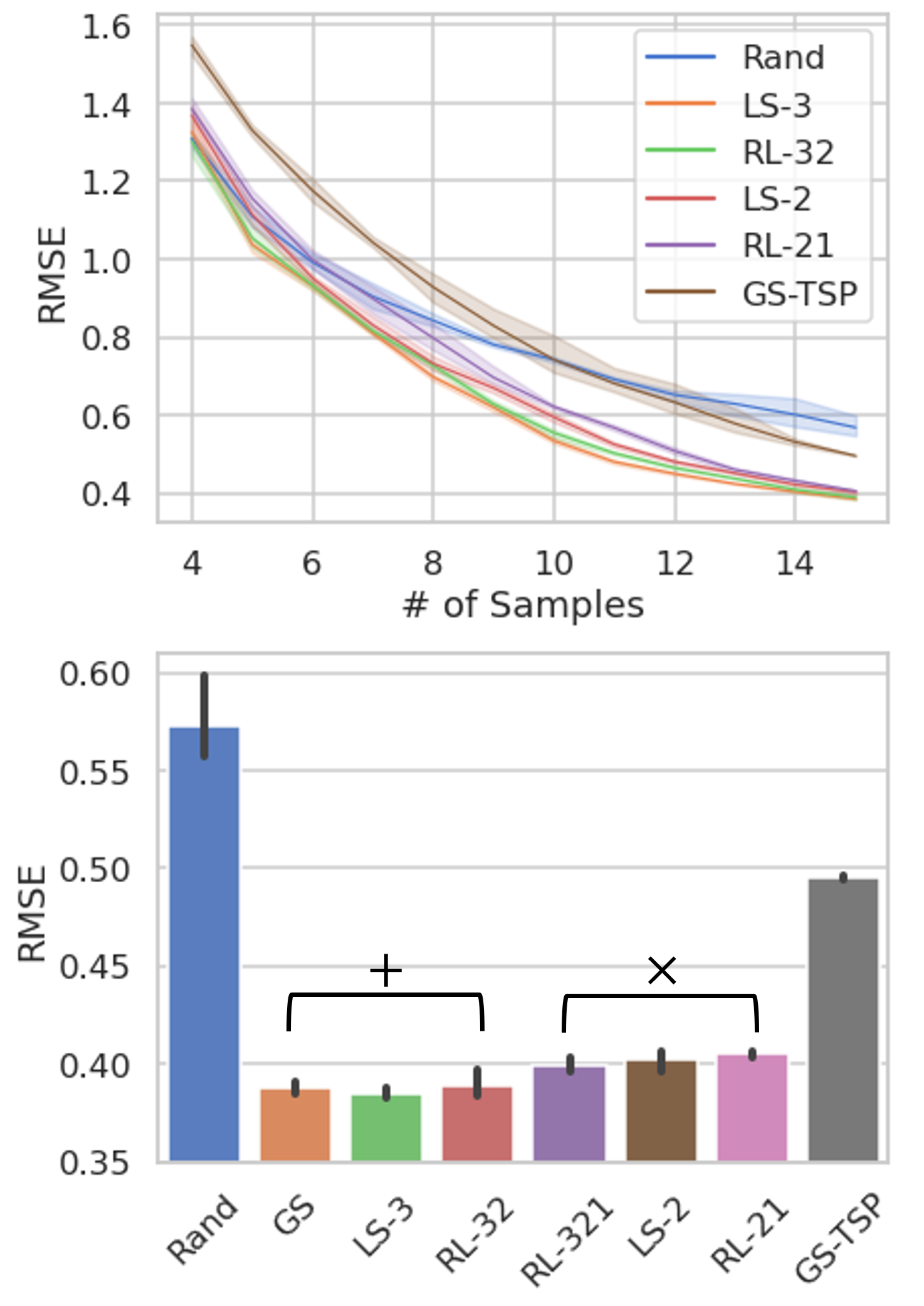}}
    \subfloat[Travel Distance]{\label{fig:travel_distance}%
        \includegraphics[width=0.495\columnwidth]{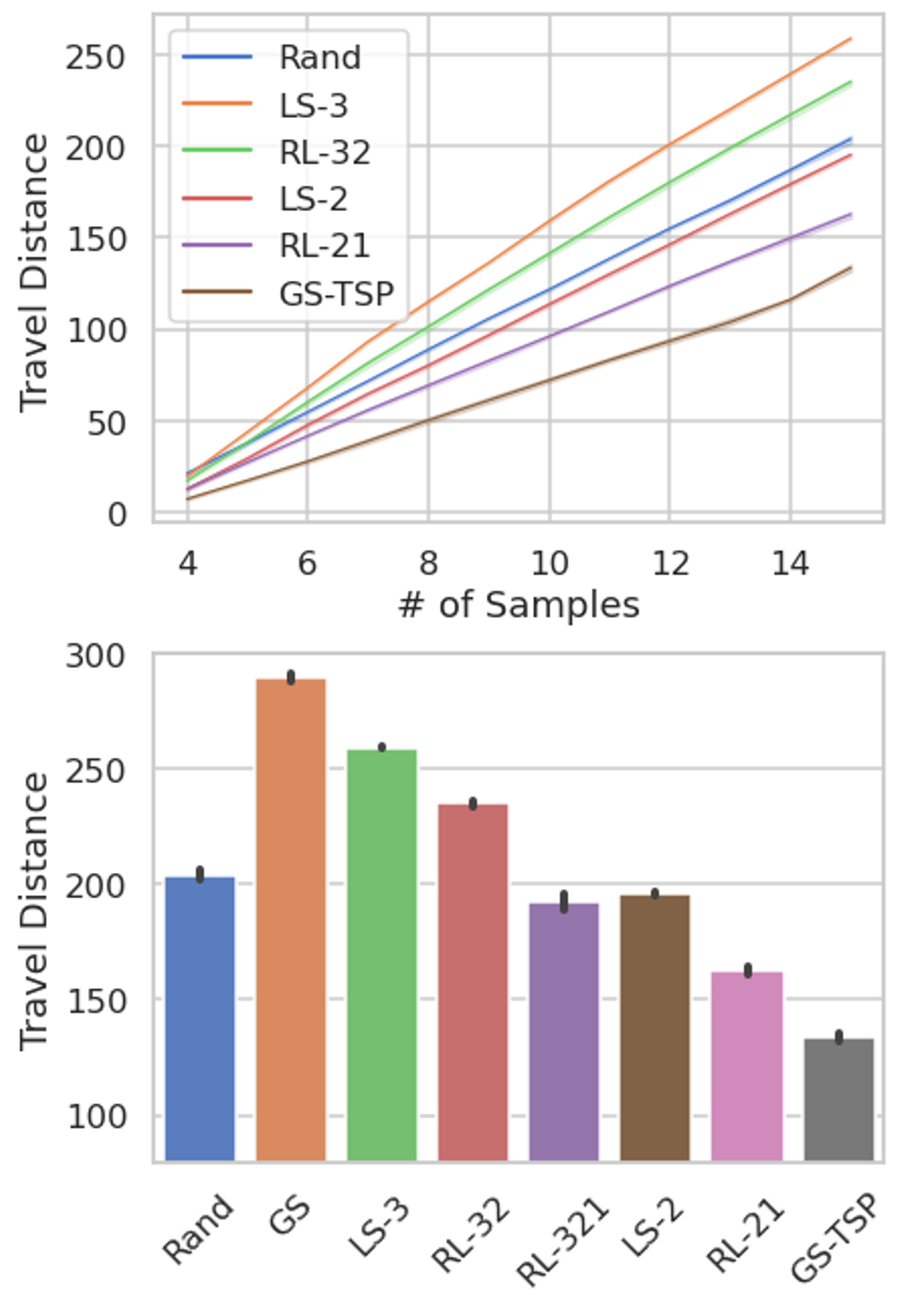}}
    \caption{
        Average prediction errors and travel distances from 
        three individual runs of each method --- the results after each new
        sample~(Top) and after the last $15$th sample~(Bottom). 
        ``$+$'' and ``$\times$''~indicate insignificant difference 
        from LS-$3$ and LS-$2$, respectively, with $p > .05$ in 
        \emph{t-test}. 
        }
    \label{fig:prediction_error_travel_distance}
\end{figure}

\section{Experiments}
\label{sec:experiments}

In this section, we show experimental results to validate our aforementioned 
hypotheses in~\autoref{sec:integration_with_tsp_solvers} and novel model 
designs in~\autoref{sec:proposed_methodology} compared to other baselines. 
Additionally, qualitative analyses are conducted with visualizations of 
adaptive selection among planning policies to discuss meaningful regularities 
of strategy choice.  

\subsection{Baselines \& Protocols}

\begin{itemize}
    \item \emph{Rand}: Random policy to choose $T$ arbitrary locations 
                    for path planning.
    \item \emph{GS}: KV-based policy with global search for the largest KV~\citep{FGDPC18}.
    \item \emph{GS-TSP}: GS with TSP solvers~\citep{FGDPC18}, for which Python 
                        TSP Solver\footnote{https://github.com/fillipe-gsm/python-tsp} 
                        is used for efficient approximation.
    \item \emph{LS-$R$}: KV-based policy with the application of local planning radius 
                        $R\times 10$ where $R\in \{2, 3\}$.
    \item \emph{RL-$K$}: Proposed RL-based controller, denoted with $K\in\{32, 21, 332\}$ 
                    indicating which LS policies are considered for action ---
                    e.g.,~RL-32 for $\{\text{LS-}3, \text{LS-}2 \}$.
\end{itemize}

As stated in~\autoref{sec:use_case}, temperature data is used with $120$~instances to test 
all models as well as an exclusive set of $120$~instances to train RL models. Also, 
$15$~samples are set to the budget~$T$ for sampling since we have found that 
further sampling offers sufficient observations for most baselines to reach
a similar, small error in final prediction.
Furthermore, first three \emph{seed} locations~$\mathcal{L}_3$  
at $((1,1), (2,2), (3,3))$ are used for all methods to learn initial 
parameters of OK as explained in~\autoref{sec:ordinary_kriging}. 
Moreover, prediction errors are measured by 
RMSE in~\autoref{eq:rmse}, and travel distances are by the sum of \emph{Euclidean distances} 
between consecutive traversed points in paths.
In particular, \emph{PyKrige}\footnote{https://pykrige.readthedocs.io} in Python
is utilized to solve the OK system in~\autoref{sec:ordinary_kriging}, 
and also, we implement DQN models for RL using 
\emph{Stable-Baselines}\footnote{https://stable-baselines3.readthedocs.io}.
\Autoref{fig:rl_learning} displays an instance of learning progress in our 
RL model performing $100$K interactions for an hour on a 
NVIDIA GeForce GTX 1080Ti GPU and a Ryzen 9 3950X CPU.

\begin{figure*}[t] \centering
    \includegraphics[width=2\columnwidth]{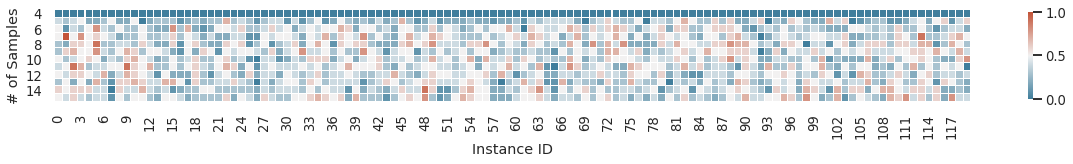}
	\caption{Average actions in RL-$21$ over $10$~independent 
    runs on test instances. Blue (red) represents choices of LS-$2$ (LS-$1$).}
    \label{fig:action_history}
\end{figure*}

\subsection{Comparative Evaluation}
\label{sec:comparative_evaluation}

\Autoref{fig:prediction_error} shows every approach presents 
a constant decline of error as more observations are collected, 
but the rate of reduction can vary. 
For instance, the random planning is the slowest in error reduction  
implying that KV obviously offers useful guidance to obtain informative 
observations reconfirming our claim in~\citep{FGDPC18}. Also, 
GS-TSP underperforms other KV methods despite its shortest travel distance 
in~\autoref{fig:travel_distance}, and our further analysis on 
potential factors are described in~\autoref{sec:problem_in_GS_TSP}.

In addition, RL-$32$ and RL-$21$ can achieve similar levels of accuracy to 
their counterparts, LS-$3$ and LS-$2$, respectively, while 
traveling significantly shorter paths as shown in~\autoref{fig:travel_distance}. 
Particularly, RL-$32$ reduces $19\%$ of total travel distance to provide
equivalent accuracy to the GS method. 
These results essentially support our hierarchical design of RL-based controller 
over lower-level strategies to maintain the maximised prediction accuracy 
from efficient navigation plans. The same could not be easily realised 
by employing a single strategy only as discussed in~\autoref{fig:tradeoff}. 
Still, the lower performance of RL-$321$ than LS-$3$ indicates the need of 
improved design to better learn with larger action spaces. 

\subsubsection{Negative Factors in GS-TSP}
\label{sec:problem_in_GS_TSP}

We first have investigated the rate of the ``TSP random samples''~(c.f.,~\autoref{sec:integration_with_tsp_solvers}) 
finally visited by the robot, and three individual runs of GS-TSP have shown 
that out of $12$~sampling locations after three initial seeds, 
$3.8$~locations belong to the TSP's initial random set in average. 
Moreover, the locations that were of the highest KV when added to the 
planning queue are eventually ranked at only top~$34\%$ at the time of visit. 
These observations empirically prove the relatively low ``utility'' of 
samples in GS-TSP, and in result, higher prediction errors when compared to 
other KV-based planners in~\autoref{fig:prediction_error}.

\subsection{Adaptive Action Selection and its Positive Impacts}
\label{sec:adaptability_of_action_selection}

\Autoref{fig:action_history} visualizes the series of average actions 
selected by RL-$21$ for each location decision in test instances. 
Most samples involve highly variable selections of LS policies depending 
on appearing state contexts in prediction though for the first sample of~$\ell_4$, 
the action is optimized to consistently execute LS-$2$ for any instance. 
This may be because starting with an observation from a larger range could 
provide a better understanding of the environmental model for future 
planning especially when our current configuration sets the \emph{seed} 
data points to be gathered only from a narrow region.

\subsubsection{Utility of Contextual Input}

Here, we design a simplistic method for action selection relying solely on the 
empirical probabilities discovered from~\autoref{fig:action_history}
so as to investigate whether this context-free approach can still achieve the equivalent 
performance to our RL model. 
$P'(a_t = \text{LS-}2)=.62$ and $P'(a_t = \text{LS-}1)=.38$  
were first calculated, 
and the average performance over $10$~separate runs of each method 
demonstrated the significant superiority of RL-$21$ in precision because it led
the error of $.400$ $(\pm.006)$ against 
$.415$ $(\pm.014)$ with $t(18)=2.976$, $p=.011$ in \emph{t-test}. 
Furthermore, the margin appears more significant at earlier steps 
(e.g.,~$p=.001$ and $p=.002$ after $13$~and $14$~samples, respectively).
In other words, our RL agent does not perform arbitrary action selections
but makes use of the contextual information provided in input, and also, 
this context-aware adaptive behavior is essential to maximise the 
overall prediction reliability when multiple plans are blended.

\section{Conclusion \& Future Work}
\label{sec:conclusion}

We have proposed a novel framework to build RL-based controllers 
to adaptively select IPP strategies based on the latest states of 
interpolation.
In a scenario of an autonomous temperature monitoring, we have
demonstrated the utility of combining multiple local-range planners 
over relying only on a single planning policy by showing
the maximised accuracy of predicted environmental model
while more efficient paths are planned for spatial sampling. 
Lastly, the dynamic policy selection has revealed that intermediate 
outcomes from interpolator can be a useful source of information to 
increase adaptability in informative planning. 

In future work, real mobile robots could be implemented to further validate 
the reliability of the RL~model under noisy environments. 
In addition, our framework could be extended to incorporate 
the travel distance explicitly into the cost function in optimization to 
mathematically satisfy resource constraints. Moreover, the transferability 
of learned parameters in neural networks could be tested between 
datasets of different geographical regions or spatial attributes. 
Similarly, generalizable learning methods could be invented to adapt 
the policy learnt from a local region to the entire field area.


{\small
    \ifx\usenatbib\undefined%
	\bibliographystyle{IEEEtran}%
    \else%
    \bibliographystyle{IEEEtranN}%
    \fi
	\bibliography{bib}
}

\end{document}